%% file: acl_latex.tex
\pdfoutput=1
\documentclass[11pt]{article}

\usepackage[]{acl}

\usepackage{times}
\usepackage{latexsym}

\usepackage[T1]{fontenc}
\usepackage[utf8]{inputenc}

\usepackage{microtype}

\usepackage{inconsolata}

\usepackage{xcolor}
\PassOptionsToPackage{dvipsnames}{xcolor}
\usepackage{multirow}
\usepackage{multicol}
\usepackage{graphicx}
\usepackage{tabu}
\usepackage{amsmath}
\usepackage{booktabs}
\usepackage{hyperref}

\usepackage{subcaption}
\usepackage{caption}
\usepackage[normalem]{ulem}
\usepackage{soul}
\usepackage[shortlabels]{enumitem}
\usepackage{array}
\usepackage{pgffor}
\usepackage{multirow}
\usepackage[normalem]{ulem}
\usepackage{array}
\usepackage{ragged2e}
\usepackage{tcolorbox}

\title{Towards Enhancing Coherence in Extractive Summarization:\\Dataset and Experiments with LLMs}

\author{Mihir Parmar$^{1*}$ \quad Hanieh Deilamsalehy$^2$ \quad Franck Dernoncourt$^2$ \\\textbf{Seunghyun Yoon}$^2$ \quad \textbf{Ryan A. Rossi}$^2$ \quad \textbf{Trung Bui}$^2$ \\\\ $^1$Arizona State University, USA \\ $^2$Adobe Research, USA \\
\small{\texttt{mparmar3@asu.edu}, \texttt{deilamsa@adobe.com}}}

\begin{document}
\maketitle
\begin{abstract}
Extractive summarization plays a pivotal role in natural language processing due to its wide-range applications in summarizing diverse content efficiently, while also being faithful to the original content. Despite significant advancement achieved in extractive summarization by Large Language Models (LLMs), these summaries frequently exhibit incoherence. An important aspect of the coherent summary is its readability for intended users. Although there have been many datasets and benchmarks proposed for creating coherent extractive summaries, none of them currently incorporate user intent to improve coherence in extractive summarization. Motivated by this, we propose a systematically created human-annotated dataset consisting of coherent summaries for five publicly available datasets and natural language user feedback, offering valuable insights into how to improve coherence in extractive summaries. We utilize this dataset for aligning LLMs through supervised fine-tuning with natural language human feedback to enhance the coherence of their generated summaries. Preliminary experiments with Falcon-40B and Llama-2-13B show significant performance improvements ($\sim 10\%$ Rouge-L) in terms of producing coherent summaries. We further utilize human feedback to benchmark results over instruction-tuned models such as FLAN-T5 which resulted in several interesting findings\footnote{Data and source code are available at \url{https://github.com/Mihir3009/Extract-AI}}.

\def\thefootnote{*}\footnotetext{Work done while interning at Adobe Research.}\def\thefootnote{\english{footnote}}

%We hope that our findings facilitate future research for improving coherence in extractive summarization
%and improving coherence in extractive summaries is under-explored

% \mihir{Comment}
% \hanieh{Comment}
% \franck{Comment}
% \trung{Comment}
% \ryan{Comment}
% \david{Comment}

\end{abstract}

\input{01_introduction}

\input{03_data_annotation}

\input{04_experiments}

\input{05_conclusion}

% Entries for the entire Anthology, followed by custom entries
\bibliography{anthology,custom}

\clearpage

\appendix

\input{appendix}

\end{document}

%% file: 01_introduction.tex
\section{Introduction}

With the increasing amount of information, the significance of automatic summarization has grown exponentially. Summarization techniques can be broadly classified into two categories: (i) Extractive, and (ii) Abstractive. The abstractive methods \cite{nallapati-etal-2016-abstractive, gupta2019abstractive} often focus on the semantic meaning of the text, giving a summary by creating a new set of sentences. However, these methods often struggle with generating ungrammatical or even nonfactual contents \cite{kryscinski-etal-2020-evaluating, zhang-etal-2022-improving-faithfulness}. In contrast, extractive methods focus on selecting meaningful phrases/sentences from the given text, giving a summary that is faithful to the original content, hence it has a range of real-world applications \cite{zhang2023extractive}. For instance, tasks such as video shortening, and legal document summarization require precision and adherence to specific details from original text, and extractive methods are more suitable for these tasks. However extractive summarization often generates summaries that lack coherence, and coherence is a crucial attribute of text summarization since it holds a significant connection to user experience. Thus, our work aims to improve coherence in extractive summarization. 

%and profoundly influences the comprehensibility of the generated or extracted summaries

\begin{figure}
    \centering
    \includegraphics[width=\linewidth]{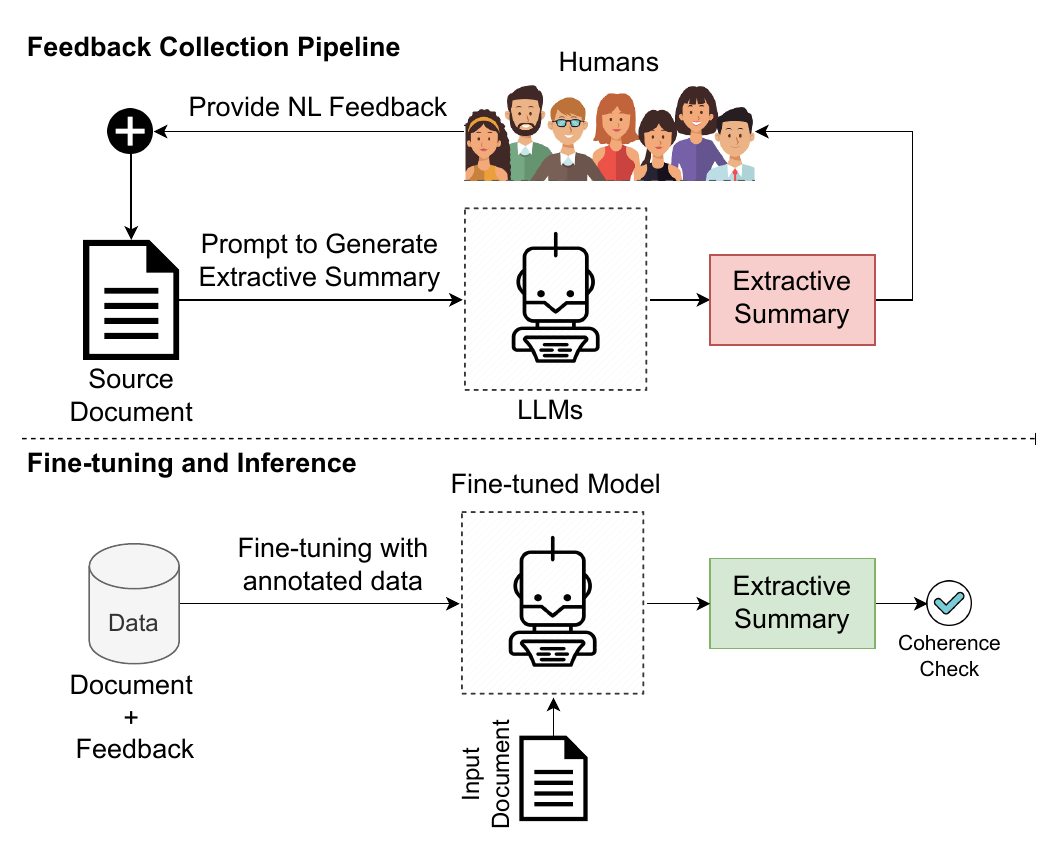}
    \caption{Schematic representation of our natural language feedback collection pipeline and aligning LLMs with provided human feedback.}
    \label{fig:teaser_figure}
\end{figure}

%\hanieh{you can also add something like even though the recent advances in LLMs have resulted in better extractive summaries (meaning that we can select important parts of the text) these summaries are sometimes hard to read and jumpy (aka lack of coherence)}.

%Coherence plays a pivotal role in creating a high-quality summary of a document.

With the advent of LLMs such as GPT-4, Llama-2 \cite{touvron2023llama}, and Falcon \cite{refinedweb}, there is a significant advancement in generating extractive summaries \cite{zhang2023extractive, stiennon2020learning}. For extractive summarization, coherence is often measured through the interconnection among sentences and ease of readability for users. Past attempts have been made to improve and quantify coherence in extractive summarization \cite{nallapati-etal-2016-abstractive, Wu2018LearningTE, jie2023enhancing}\footnote{Detailed related work is presented in App. \ref{app:related_work}}, however, these attempts do not consider user-specific intent (i.e., ease of readability while preserving important information). Thus, we approach the concept of coherence through the lens of user-specific intent (Figure \ref{fig:teaser_figure}). To this end, we propose a comprehensive dataset with a systematic collection of natural language feedback to improve coherence in model-generated summaries, and human-annotated extractive coherent summaries. To the best of the authors' knowledge, this dataset represents the initial effort to align the coherence in a summary with user intent.

To develop the proposed dataset, we hire expert annotators to accurately annotate data for our task. For the annotation, the objective is two-fold: (1) to create a coherent summary by extracting important sentences from a source document that effectively captures the key aspects of the document, and (2) to provide feedback (i.e, natural language explanations) on the steps to go from the model summary to the gold coherent summary. We annotate this data across five categories: News, Debate, TV Show, Meeting, and Dialogue. Our annotation process consists of three phases (detailed discussion in \textsection \ref{sec:data_annotation}). Each data instance collected in our dataset consists of \textit{<Source text, Initial model summary, Feedback, Gold coherent summary, Scores>} elements. 

%We utilize this collected data for aligning LLMs to generate more coherent summaries.

We utilize the proposed dataset for aligning widely used open-source LLMs to generate more coherent extractive summaries via supervised fine-tuning: (i) two decoder-only models, i.e., Falcon-40B and Llama-2-13B, and (ii) three encoder+decoder models, i.e., FLAN-T5, Tk-Instruct, and T5. We develop a baseline and propose two different supervised fine-tuning strategies with human feedback (details are presented in \textsection \ref{sec:experiments}). We measure the performance in terms of Rouge-L. Rouge-L assesses the syntactic and semantic similarity between the generated and the gold coherent summary, indicating their proximity. We also provide human judgments in terms of the coherence of generated summaries by baseline and proposed approach. Experimental results reveal that the proposed models show absolute improvement of $\sim 10\%$ Rouge-L over baselines. Furthermore, human evaluation shows a preference for extractive summaries from our approach, often rating them as more coherent. This indicates that aligning the model with user feedback improves coherence. Furthermore, a thorough analysis of the results reveals several interesting findings. We hope that our findings facilitate future research for improving coherence in extractive summarization.

%% file: 03_data_annotation.tex
\section{Data Collection}
\label{sec:data_annotation}

% In this section, we provide a detailed description of 
Our annotation process consists of three phases. First, we randomly select a source text for annotation across five different categories from publicly available datasets. Second, we prompt a large language model to create coherent summaries for selected source text. Finally, we hire expert annotators to review generated summaries and provide natural language feedback/explanations to improve coherence in generated summaries. 
%Furthermore, we present a quantitative analysis of our data.

\subsection{Source Datasets}

Our comprehensive annotated dataset consists of five different categories: News, Debate, TV Show, Meeting, and Dialogue. We carefully curated data for each category by randomly selecting 200 instances from publicly available datasets. In particular, we exclusively utilize the input/source text for annotation purposes from all of these datasets. We leverage CNN/DM dataset \cite{nallapati-etal-2016-abstractive} for news, DebateSum \cite{roush-balaji-2020-debatesum} for Debate, TVQA \cite{lei-etal-2018-tvqa} for TV Show, MeetingBank \cite{hu-etal-2023-meetingbank} for Meeting, and DialogueSum \cite{chen-etal-2021-dialogsum} for Dialogue category. Further details are presented in App. \ref{app:datasets}.

\subsection{Coherent Summary Generation}
\label{subsec:summary}

The objective is to generate an extractive summary, where the model is prompted to select the most suitable sentences from the document for coherent summarization. Thus, we formulate an extractive summarization task as selecting sentences from a given document to produce coherent summaries. Let us consider document $\mathcal{D}$. We first divide $\mathcal{D}$ at the sentence level and create set $\mathcal{D}_s = \{s_1, s_2, ..., s_n\}$, where $s_i$ denotes the $i^{th}$ sentence from $\mathcal{D}$. To create numbered sentences from the document, we use the NLTK library\footnote{\url{https://www.nltk.org/api/nltk.tokenize.html}}. Now, we prompt ($p$) the Falcon-40B-Instruct model (denoted as $\mathcal{M}$) to produce a coherent summary from the source text provided as $\mathcal{D}_s$. To accomplish this, we employ a 1-shot prompting approach (prompt is presented in the App. \ref{app:prompt}). Formally, we present our task as $\mathcal{M}(p, \mathcal{D}_s) = C_s$, indicates that the task for $\mathcal{M}$ is to produce coherent summary (denoted as $C_s$) by selecting sentences from $\mathcal{D}_s$ given $p$. 

%An example of a 1-shot prompt used for producing this initial model summary .

% During this stage, we prompt the Falcon-40B-Instruct model to produce a coherent summary from the source text. The primary objective is to generate an extractive summary, where the model is prompted to select the most suitable sentences from the document for summarization. To accomplish this, we employ a 1-shot learning approach. We also present the source text in terms of numbered sentences, as demonstrated in the prompt below. To create numbered sentences from the document, we use the NLTK library\footnote{\url{https://www.nltk.org/api/nltk.tokenize.html}}. 

%\hanieh{Maybe you can formulate the task a bit here and write some sort of a formula including sentence numbers and sentences and the summarization task and then show an example. The reason is that the traditional summarization task might not have the sentences as unit and does not include sentence numbers.}

\subsection{Annotation Process}

\begin{figure}[h]
    \centering
    \includegraphics[width=0.55\linewidth]{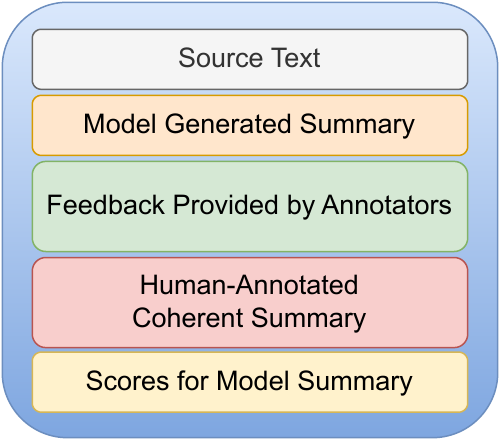}
    \caption{Illustration of annotated instance}
    \label{fig:annotation}
\end{figure}

We use the Upwork platform to hire expert annotators to annotate our dataset. We initiated a pilot project involving 25 annotators having a strong background and fluency in the English language. Evaluating their performance during the pilot phase, we subsequently hired 10 proficient annotation experts to carry out the final annotations. Annotators are provided with task instructions, source text, and model summary (generated in \textsection \ref{subsec:summary}). They are expected to produce a coherent summary based on the provided source text by selecting sentences/phrases from the document and provide feedback on the steps to go from the model summary to the gold coherent summary (annotated by them). Each source text is annotated by 3 different annotators. Along with that, they need to rate the model summary based on three criteria (i.e., Relevance, Coherence, and Consistency) on a Likert scale of 1-5, motivated by \citet{fabbri-etal-2021-summeval}. A annotated data instances consist of five elements as illustrated in Figure \ref{fig:annotation}. A detailed example and further annotator details are presented in App. \ref{app:example}.

%for each element
%\footnote{\url{https://www.upwork.com/}}

\begin{figure*}
\centering
\begin{subfigure}{0.5\textwidth}
  \centering
  \includegraphics[width=\linewidth]{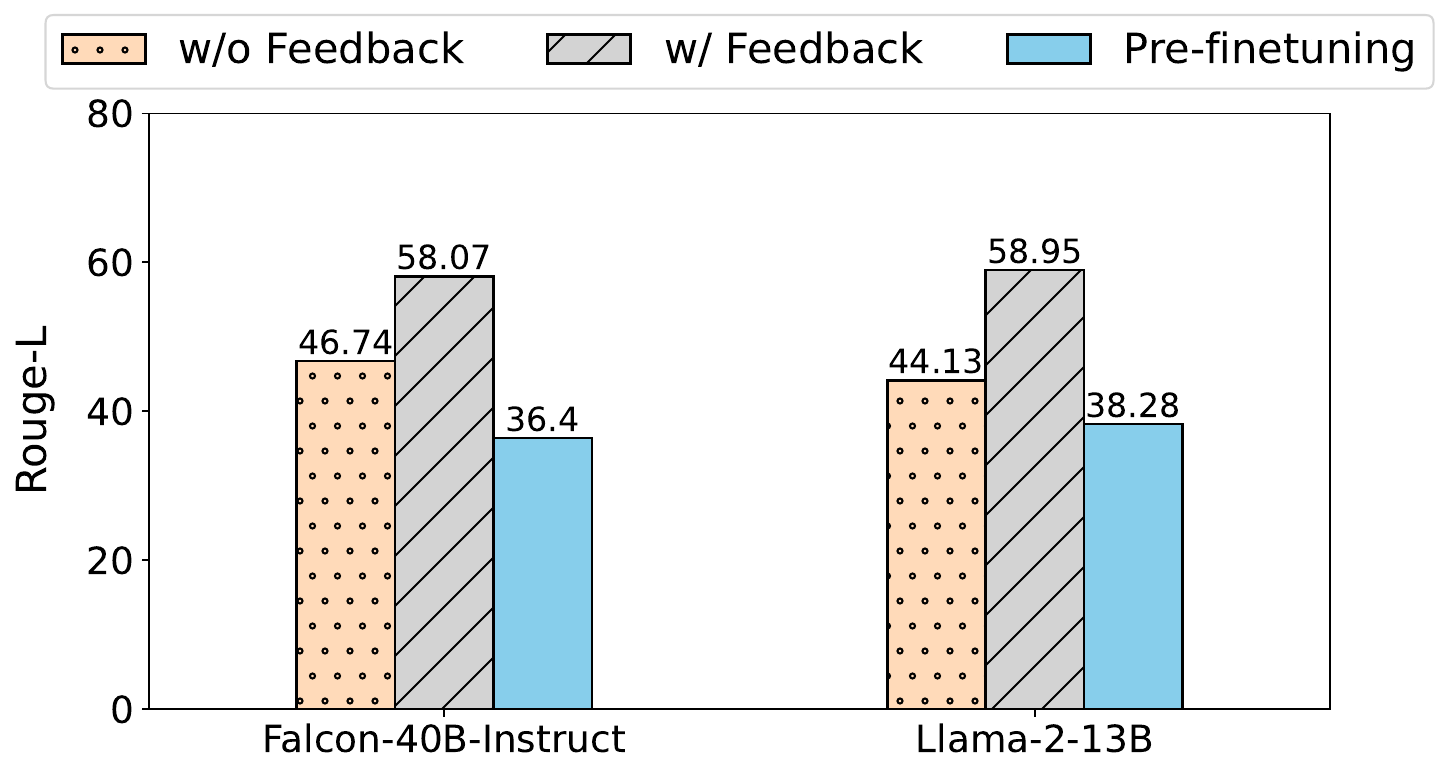}
  \caption{}
  \label{fig:dec_only}
\end{subfigure}%
\begin{subfigure}{0.5\textwidth}
  \centering
  \includegraphics[width=\linewidth]{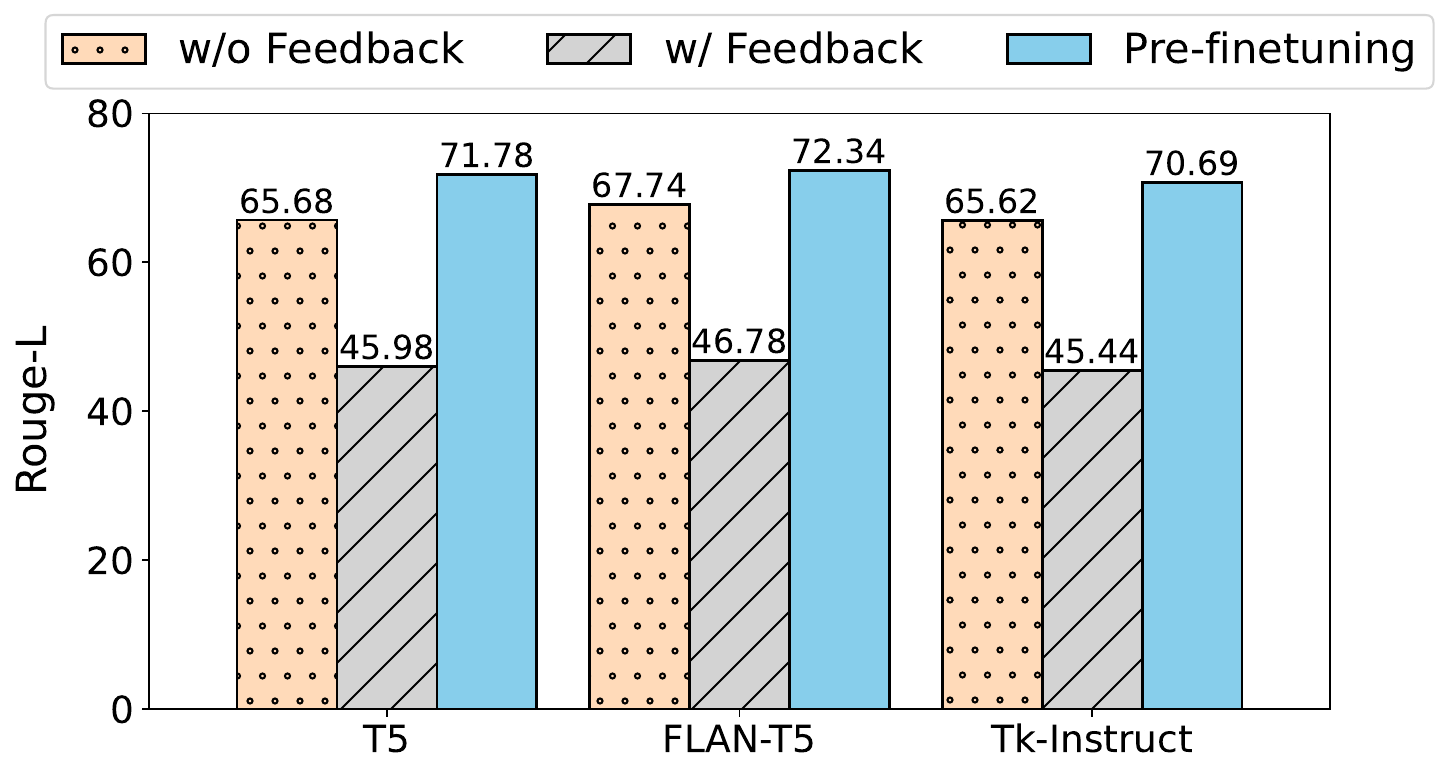}
  \caption{}
  \label{fig:enc_dec_only}
\end{subfigure}
\caption{Performance of (a) Dec. only model, and (b) Enc. + Dec. Model on our proposed dataset.}
\label{fig:main_results}
\end{figure*}

\paragraph{Source text} is the document provided to annotators which falls under one of five categories.

\paragraph{Model-generated summary} The summary generated in \textsection \ref{subsec:summary} is provided to annotators.

%corresponding to each source text

\paragraph{Coherent Summary} is generated by annotators from the given source document.

\paragraph{Feedback} is a natural language explanation provided by annotators to improve coherence in the model summary and achieve a coherent summary generated by them.

\paragraph{Scores} Annotators score the model-generated summary to measure the three different aspects: (i) Relevance: measure the selection of important content (key points) from the source, and the summary should include only important information from the source document; (ii) Coherence: measure the collective quality of all sentences, and the summary should be well-structured and well-organized; and (iii) Consistency: measure the factual consistency of the summary that contains only statements that are entailed by the source document.

\subsection{Quantitative Analysis}

Annotators have annotated a total of 1000 unique samples and each sample is annotated by three different annotators with the inter-annotator agreement of 0.659 (details in App. \ref{app:inter_agree}). For each document category, 200 samples are annotated. After all annotations, the average scores for model summary are: (1) Relevance: 3.81, (2) Coherence: 3.46, (3) Consistency: 4.09. Here, coherence is low for the model-generated summary which suggests that improving coherence is essential task.

%For the annotation, the objective is two-fold: (1) to create a coherent summary extracting important sentences/phrases from a document that effectively captures the key aspects of the document, and (2) to provide feedback on the steps to go from the model summary (generated in \textsection \ref{subsec:summary}) to the coherent summary from step (1).

%% file: 04_experiments.tex
\section{Experiments and Results}
\label{sec:experiments}

% In this section, we provide comprehensive information regarding our experimental setup and the obtained results. Additionally, we delve into a detailed analysis of the results.

\subsection{Experimental Setup}

\paragraph{Models} We perform experiments with five different models with two architecture families: (i) two Decoder (Dec.) only open-source LLMs (Falcon-40B, and Llama-2-7B), and (ii) three Encoder (Enc.) + Decoder (Dec.) models (T5-large, and two instruction-tuned models, FLAN-T5-large and Tk-Instruct-large). In experiments, Dec. only models are fine-tuned using Low-Rank Adaptation (LoRA) \cite{hu2021lora}, and Enc.+Dec. models are fine-tuned using full-parametric training. We employ three different strategies to fine-tune these models.

\paragraph{Baseline} fine-tuning model on \textit{<Source text>} as input and \textit{<Coherent Summary>} as output. 

%In particular, the baseline approach utilizes the document as input and the coherent summary as the output label.

\paragraph{w/ Feedback} fine-tuning model on \textit{<Source text, Initial model summary, Feedback>} as input and \textit{<Coherent Summary>} as output. 

%this proposed method involved incorporating a more comprehensive input, encompassing the document, model-generated summary, and user feedback.

\paragraph{Pre-finetuning} First, we fine-tune the models on \textit{<Source text>} as input and \textit{<feedback>} as the output. Subsequently, we execute supervised fine-tuning by employing \textit{<Source text>} as the input and \textit{<Coherent Summary>} as the output on the pre-finetuned model.

Our approaches reflect an effort to refine the models' coherence by leveraging feedback and user-driven insights during the fine-tuning. 
We fine-tune the model to generate sentences as a summary (format of the coherent summary is shown in Table \ref{tab:example_annotation}) which ensures the extractive nature of generated summaries.
The dataset is randomly divided into train (80\%), and test (20\%) sets. For comparability, we use the same hyperparameter settings for all runs: trained for 3 epochs, with a batch size of 16 and an initial learning rate of 5e-5. All experiments were conducted on A100 NVIDIA GPUs.
% The dataset is randomly divided into train (80\%), and test (20\%) sets. For comparability, we use the same hyperparameter settings for all runs: training is run for 3 epochs, with a batch size of 16 and an initial learning rate of 5e-5. All experiments were conducted on A100 NVIDIA GPUs.

\paragraph{Metric} 
We use Rouge-L \cite{lin-2004-rouge} to evaluate model performance by measuring the similarity between the generated summary and the gold standard coherent summary. Our assessment is based on how closely the model summary resembles this gold standard, indicating coherence similarity. To supplement this objective measure, we also perform human evaluations of the generated summaries.

%and G-Eval \cite{liu2023g}

%We utilize Rouge-L \cite{lin-2004-rouge} as an evaluation metric for assessing model performance. Rouge-L quantifies the syntactic and semantic similarity between the generated summary and the gold coherent summary. The foundation of our evaluation lies in the fact that the reference summary we employ as the gold standard represents a coherent summary, and the proximity of the generated model summary to this gold standard is indicative of their coherence similarity. To supplement this objective measure, we also perform human evaluations of the generated summaries.

\subsection{Results and Analysis}

Here, we compare the baselines and proposed methods despite different fine-tuning approach since the inference is consistent: \textit{<Source text>} is input, and \textit{<Coherent Summary>} is output. Models do not have access to feedback during inference.

\paragraph{Effect of Feedback on Dec. only models} 
% Figure \ref{fig:dec_only} illustrates the Rouge-L scores for Falcon-40B-Instruct and Llama-2-13B, comparing both baseline and proposed methods. The results distinctly showcase the superiority of the proposed methods, which involve fine-tuning with user feedback. Specifically, the Falcon surpasses the baseline by 11.33\%, while the Llama model outperforms the baseline by 14.82\%. However, the performance of both models significantly declines when subjected to pre-finetuning with feedback data. The primary objective of pre-finetuning with feedback is to integrate the feedback knowledge into the model's parameters. As we fine-tune both models using LoRA, wherein only the adaptation layer is updated, there is a noticeable drop in performance during pre-finetuning. Conversely, the efficacy of pre-finetuning becomes evident in the context of full-parametric training, as depicted in Figure \ref{fig:enc_dec_only}. 

Figure \ref{fig:dec_only} shows the Rouge-L scores for Falcon-40B-Instruct and Llama-2-13B, comparing baseline and proposed methods. The proposed methods, involving fine-tuning with user feedback, clearly outperform the baselines: Falcon improves by 11.33\%, and Llama by 14.82\%. However, both models' performance drops significantly during pre-finetuning with feedback data. This pre-finetuning aims to integrate feedback knowledge into the model's parameters. When fine-tuning with LoRA, updating only the adaptation layer, performance decreases during pre-finetuning. However, the efficacy of pre-finetuning becomes evident with full-parametric training, as shown in Figure \ref{fig:enc_dec_only}.

% \begin{figure}[h]
%     \centering
%     \includegraphics[width=\linewidth]{images/results.pdf}
%     \caption{LoRA Fine-tuning: Falcon and Llama-2}
%     \label{fig:result_1}
% \end{figure}

\paragraph{Effect of Feedback on Enc. + Dec. models}
Figure \ref{fig:enc_dec_only} represents the Rouge-L scores for FLAN-T5, Tk-Instruct, and T5, comparing both baseline and proposed methods. From the results, it becomes evident that directly fine-tuning with user feedback doesn't enhance the performance of these models as shown with Dec. only models. Conversely, adopting a pre-finetuning enhances the performance of these models significantly (further discussion in App. \ref{app:human_eval}). Figure \ref{fig:enc_dec_only} shows that pre-finetuning leads to improved performance, with the T5, FLAN-T5, and Tk-Instruct models surpassing baseline by 6.1\%, 4.6\%, and 5.07\%, respectively.

%The data in Table \ref{tab:enc_dec_results} corroborates that pre-finetuning significantly contributes to improving performance.

% \input{tables/enc_dec_results}

% \paragraph{Qualitative Analysis of Generated Summries}
% \mihir{Add analysis of how generated summaries are better and what they are still lacking}

\paragraph{Human Evaluation}
We aim to examine the correlation between human judgments and Rouge-L. To this end, we conduct a case study involving human evaluation. We asked three independent human evaluators (graduate student volunteers) to assess the summaries (50 randomly selected from the test set). Each evaluator was asked to choose their preferred summary from three options: (1) the model summary (provided during annotations), (2) Llama-2 (w/o feedback), and (3) Llama-2 (w/ feedback). Additionally, they were asked to rate each summary's coherence on a Likert scale ranging from 1 (incoherent) to 5 (perfectly coherent). 
We calculate the inter-annotator agreement based on their choice of preferred summary. Since coherence is very subjective to annotators, we found 0.513 inter-annotator agreement (measured with raw/observed agreement) between three different annotators.

\begin{figure}[h]
    \centering
    \includegraphics[width=\linewidth]{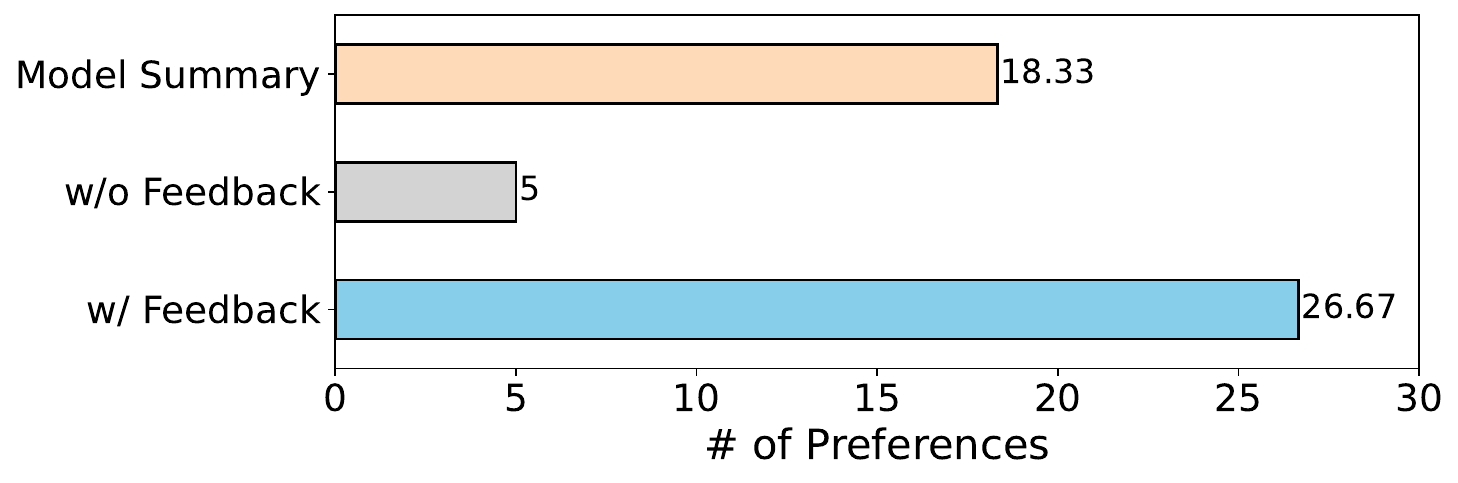}
    \caption{Average number of preferences across three evaluators.}
    \label{fig:human_eval}
\end{figure}

Figure \ref{fig:human_eval} shows the results for an average number of preferences across three evaluators, and the average coherence score is 3.45, 2.29, and 3.53 for model summary, Llama-2 (w/o feedback), and Llama-2 (w/ feedback), respectively. The results revealed that, on average, the evaluators favored the summary from Llama-2 (w/ feedback), which also received the highest average coherence score. These findings are consistent with and further corroborated by the results presented in Figure \ref{fig:dec_only}. This further supports the findings presented in the paper using Rouge-L. 

%presented in App. \ref{app:human_eval}.
% with 0.785 inter-annotator agreement (measured with raw/observed agreement)
%of summaries generated by Llama-2. The results are 

%% file: 05_conclusion.tex
\section{Conclusions}
This paper introduced a comprehensive dataset designed to improve coherence in extractive summarization while integrating natural language feedback from human users across five different categories. Utilizing this dataset, we conducted evaluations using various LLMs, and initial experimental outcomes demonstrate an enhancement in model performance, with $\sim10\%$ improvement in coherence achieved through fine-tuning with human feedback. Moreover, our analysis highlights the potential for performance advancements in instruction-tuned models through pre-finetuning based on user feedback. We believe that both the dataset and the findings derived from this work will serve as valuable tools for future research in this direction. 

\section*{Limitations}

Though we evaluated our approach on a widely-used range of LLMs including Falcon-40B and LLaMa-2-7B, this study can also be extended to other LLMs. To improve the utilization of human feedback collected in our dataset, development of advanced methods such as iterative feedback loops and dynamic feedback during both training and inference stages can be interesting future research direction. Since manual annotation of feedback is time-consuming and laborious, exploration of automated methods for feedback generation using smaller-scale supervised learning or LLMs is necessary. Additionally, we hope to expand our analysis to include the most recent LLMs such as GPT-4 and ChatGPT on our proposed dataset. We also note that this research is limited to the English language and can be extended to multilingual scenarios for improving coherence in extractive summarization.

\section*{Ethics Statement}

We have used AI assistants (Grammarly and ChatGPT) to address the grammatical errors and rephrase the sentences.

%% file: appendix.tex
\section{Prompt}
\label{app:prompt}

In this section, we provide an example of a 1-shot prompt used in \textsection \ref{subsec:summary}. The prompt consists of the task definition, one example, and an input instance.

\begin{tcolorbox}[colback=blue!5!white]

\begin{tcolorbox}[colframe=blue!50!black,title=Task]

{\small

You are an extractive summarizer. You are presented with a document. The document is a collection of sentences and each sentence is numbered with sentence ids. Understand the given document and create a meaningful summary by picking sentences from the document. Please list the sentence IDs as output so that sentences corresponding to the generated IDs summarize the document coherently. }

\end{tcolorbox}

\begin{tcolorbox}[colframe=blue!50!black,title=Example]
{\small
Learn from the below example: 

\textbf{Document:}

1. Olympic gold medallist Jessica Ennis-Hill has confirmed she will return to competition in London this July following her break from athletics to become a mother. 

2. Ennis-Hill provided one of London 2012's most captivating storylines by surging to heptathlon gold, and the Sheffield-born star will return to the Olympic Stadium three years on to compete in the Sainsbury's Anniversary Games. 

3. The 29-year-old has not competed since the same event in 2013 and gave birth to her son, Reggie, last summer. 

.

.

.

13. Ennis-Hill will take part in the two-day meeting on July 24 and 25, with the Sainsbury's IPC Athletics Grand Prix Final taking place on July 26. 

14. Ennis-Hill added: 'The 2012 Olympics were an incredible experience for me and it will be very special to step out on that track again. 

15. It will be amazing to compete in front of all our British fans who I am sure will have their own memories of the London Games too. 

Summary: <s> [2, 5, 6, 11, 12, 15]}
\end{tcolorbox}

\begin{tcolorbox}[colframe=blue!50!black,title=Input]

{\small

Document: [source text] 

Please Create a concise summary using as few sentences as possible. 

Summary: <s> }
\end{tcolorbox}
\end{tcolorbox}

The example given in this prompt is annotated by the authors where we reviewed the document and chose specific sentence IDs to create a coherent summary.

\section{Related Work}
\label{app:related_work}

There are some past attempts that have been made to improve coherence in extractive summarization. \citet{christensen-etal-2013-towards} proposed a G-FLOW, a joint model for selection and ordering sentences that balances coherence for multi-document extractive summarization. After that, \citet{Parveen2015IntegratingIN} proposed a graph-based method for extractive single-document summarization that considers importance, non-redundancy, and local coherence simultaneously. In addition, \citet{kurisinkel2015readable} introduced A multi-document summarization method that ensures content coverage, sentence ordering, topical coherence, topical order, and inter-sentence structural relationships using a Local Coherent Unit (LCU). Following this, \citet{j-kurisinkel-etal-2016-non} proposed scoring-based function to identify the discourse structure which provides the context for the creation of a sentence for generating comprehensible summaries. Furthermore, \citet{Wu2018LearningTE} utilized reinforcement learning to extract a coherent summary, and \citet{Abdolahi2019TextualCI} enhanced coherence in extractive document summarization through a greedy approach and word vectors. In addition, \citet{Jie2023EnhancingCO} introduced two strategies, including pre-trained converting models (model-based) and converting matrices (MAT-based) that merge sentence representations to improve coherence. With the emergence of LLMs, \citet{zhang2023benchmarking} attempted to analyze the performance of GPT-3 with different prompting for generating coherent summaries. Differing from these existing efforts, we approach the concept of coherence within summaries through the lens of user-specific intent. 

%\hanieh{I think this section needs expanding and organization. Maybe describe each work a bit more and categorize the previos methods a bit rahter than listing them.}

\section{Datasets}
\label{app:datasets}

In this section, we discuss more details about publicly available datasets used for developing our proposed benchmark.

\paragraph{CNN/DM} The CNN / DailyMail Dataset is an English-language dataset containing just over 300k unique news articles as written by journalists at CNN and the Daily Mail \cite{nallapati-etal-2016-abstractive}. We utilize randomly selected 200 news articles from this dataset for our annotations. 

%\hanieh{since you have sampled these datasets randomly, do you have a poionter to connect the sampled subsets to the original dataset?}

\paragraph{DebateSum} DebateSum is constructed from evidence related to annual policy debate resolutions \cite{roush-balaji-2020-debatesum}, each averaging around 560 words. As DebateSum spans seven years of content, it encompasses seven distinct resolutions. For our annotations, we randomly selected 200 resolution plans from this dataset.

\paragraph{TVQA} TVQA is a large-scale video QA dataset based on 6 popular TV shows (Friends, The Big Bang Theory, How I Met Your Mother, House M.D., Grey's Anatomy, and Castle) \cite{lei-etal-2018-tvqa}. From this dataset, we utilize subtitles-based dialogues as source text for our annotation.

\paragraph{MeetingBank} MeetingBank is a benchmark dataset created by the city councils of 6 major U.S. cities to supplement existing datasets. It contains 1,366 meetings with over 3,579 hours of video, as well as transcripts, PDF documents of meeting minutes, agenda, and other metadata \cite{hu-etal-2023-meetingbank}. From this dataset, we utilize transcripts as source text for our annotation.

\paragraph{DialogueSum} DialogSum is a large-scale dialogue summarization dataset, consisting of 13,460 dialogues with corresponding manually labeled summaries and topics \cite{chen-etal-2021-dialogsum}. We utilize randomly selected 200 dialogues from this dataset for our annotations.

\section{Example of Annotated Instance}
\label{app:example}

In this section, we provide an example of an annotated data instance from the News category in Table \ref{tab:example_annotation}. This instance provides an illustrative example of how the whole dataset is collected. We also conduct analysis of the collected data focusing on how improving coherence affects the length of summaries, offering insights into the impact on the length of summaries. We observed that the average lengths of the original documents, model-generated summaries, and coherently annotated summaries are 24.89, 17.99, and 11.95 sentences, respectively. These findings suggest that annotators often removed sentences to enhance the coherence of the summaries during the annotation process.

\input{tables/annotator_details}

\subsection{Annotator Details} Our annotators consist of contractors hired through Upwork. Annotation of each data instance paid \$3 and could be completed within 20 minutes, compensating an annotator with an average pay of \$15/hour. The final annotation process took around time of $\sim15$ days and cost of $\sim\$10k$. Overall, we collected a total of 1000 unique samples, and the dataset was randomly partitioned into training (80\%), and test (20\%) sets. We also provide the final 10 annotators' demographic data in terms of their nationality in Table \ref{tab:annotators}.

\subsection{Calculation of Inter-annotator Agreement}
\label{app:inter_agree}
To calculate the inter-annotator agreement using ROUGE for three annotators, we focused on the ROUGE-L metric, which measures the longest common subsequence between summaries. Since the extractive summaries they have annotated are selections of sentences from the article, it makes sense to use ROUGE-L to capture the structural similarity of their selections. For each document, we computed the ROUGE-L score for every possible pair of annotators, capturing the consistency of their sentence selections. By averaging these pairwise ROUGE-L scores across all documents, we obtained an overall agreement score that reflects how closely the annotators' summaries align in terms of content and structure. This approach provides a quantitative measure of agreement that highlights the consistency among annotators in annotating the extractive summaries.

\section{Extended Discussion on Analysis}
\label{app:human_eval}
% \label{app:ext_discussion}

\paragraph{Performance of encoder-decoder \textit{vs.} decoder-only models} The observed differences in the impact of feedback on encoder-decoder models \textit{vs.} decoder-only models can be attributed to pre-training methodologies for both types of models. Encoder-Decoder models (e.g., T5, FLAN-T5) are pre-trained using a sequence-to-sequence framework, where the encoder processes the input text and the decoder generates the output text \cite{raffel2020exploring}. Decoder-only models (e.g., Falcon-40B, Llama-2) are pre-trained using a left-to-right autoregressive approach, predicting the next token based on the preceding tokens \cite{radford2019language}. When models are fine-tuned on <Source text, Initial model summary, Feedback>, decoder-only models benefit more compared to encoder-decoder models because the feedback helps them align their sequential generation process more closely with human corrections. The pre-finetuning approach involves an intermediate step where models are first fine-tuned on <Source text> as input and <feedback> as the output. For encoder-decoder models, this step helps integrate feedback more effectively into their bidirectional context understanding, leading to significant improvements. For decoder-only models, this approach does not always yield better results as they benefit more directly from feedback fine-tuning. In summary, the differential impact of feedback on encoder-decoder and decoder-only models can be attributed to their respective pre-training objectives.

% \paragraph{Human Evaluation}
% We asked three independent human evaluators (graduate student volunteers) to assess the summaries (50 randomly selected from the test set). Each evaluator was asked to choose their preferred summary from three options: (1) the model summary (provided during annotations), (2) Llama-2 (w/o feedback), and (3) Llama-2 (w/ feedback). Additionally, they were asked to rate each summary's coherence on a Likert scale ranging from 1 (incoherent) to 5 (perfectly coherent). 
% We calculate the inter-annotator agreement based on their choice of preferred summary. Since coherence is very subjective to annotators, we found 0.513 inter-annotator agreement (measured with raw/observed agreement) between three different annotators.

% \begin{figure}[h]
%     \centering
%     \includegraphics[width=\linewidth]{images/human_eval.pdf}
%     \caption{Average number of preferences across three evaluators.}
%     \label{fig:human_eval}
% \end{figure}

% Figure \ref{fig:human_eval} shows the results for an average number of preferences across three evaluators, and the average coherence score is 3.45, 2.29, and 3.53 for model summary, Llama-2 (w/o feedback), and Llama-2 (w/ feedback), respectively. The results revealed that, on average, the evaluators favored the summary from Llama-2 (w/ feedback), which also received the highest average coherence score. These findings are consistent with and further corroborated by the results presented in Figure \ref{fig:dec_only}. This further supports the findings presented in the paper using Rouge-L.

\input{tables/example_annotation} 

% \paragraph{G-Eval}

%% file: tables/annotator_details.tex
\begin{table}[ht]
\centering
\resizebox{0.6\linewidth}{!}{
\begin{tabular}{c|c}
\toprule
\textbf{Nationality} & \textbf{\# of Annotators} \\ \midrule
India                & 3                         \\
Philippines          & 3                         \\ 
Venezuela            & 1                         \\ 
Pakistan             & 1                         \\ 
Macedonia            & 1                         \\ 
Kenya                & 1                         \\ \bottomrule
\end{tabular}
}
\caption{Demographic details of annotators}
\label{tab:annotators}
\end{table}

%% file: tables/example_annotation.tex
\begin{table*}
\centering
\resizebox{\linewidth}{!}{
\begin{tabular}{p{1.4\linewidth}}
\toprule
\textbf{Document:}\\If anyone won this debate it was the women. Their less choreographed style of body language gave the impression we were listening to real messages from real people rather than watching spin doctors’ puppets performing. Overall I’m sure Miliband’s coaching team will be patting themselves on the back and .............                                                                                                                                                                                                                                                                                                                                                                                                                                                                                                                                                                                                                                                                                                                                                                                                    \\ \midrule
\textbf{Model Summary:}\\Their less choreographed style of body language gave the impression we were listening to real messages from real people rather than watching spin doctors’ puppets performing. Nicola Sturgeon (pictured) is a smiling assassin,  ........                                                                                                                                                                                                                                                                                                                                                                                                                                                                                                                                                                                                                                                                                                                                                                                                                                                                                    \\ \midrule
\textbf{Coherent Summary:} \\ Sent. 1: If anyone won this debate it was the women.  \\ Sent. 2: Their less choreographed style of body language gave the impression we were listening to real messages from real people rather than watching spin doctors’ puppets performing. \\ Sent. 3: In his après-Paxman mode, David Cameron (pictured) was looking serious and oozing leadership charisma . \\ \\  .....                                                                                                                                                                                                                                                                                                                                                                                                                                                                                                                                                                                                                                                                                                                                        \\ \midrule
\textbf{Feedback:}\\Sent. 1: If anyone won this debate it was the women.  \\ Feedback 1: Add this sentence to give an idea what the summary is all about. \\ . \\ . \\ Sent. 6: Clegg is a good speaker but his performance was vintage, ie a complete re-run of his 2010 routine.  \\ Feedback 6: Add this sentence in the model summary to provide information about the speaker. \\ . \\ . \\ Sent. 9: He took enough pops at Cameron and waved his arm enough in that direction to signal an official end to the relationship that began in the Rose Garden but he looked more congruent agreeing with Cameron or fielding criticism as a double act than he did turning on him, which looked rather panto.   \\ Feedback 9: Add this sentence in the model summary as a supporting detail to the previous sentence. \\ \midrule

\textbf{Scores:}
\\Relevance: 4                                                                                                                                                           \\Coherence: 3                                                                                                                                                           \\ Consistency: 5   \\ \bottomrule
\end{tabular}
}
\caption{Illustrative example of annotated instance. Certain text is redacted due to space constraints.}
\label{tab:example_annotation}
\end{table*}